\documentclass[letterpaper, 10 pt, conference]{ieeeconf}  

\IEEEoverridecommandlockouts                              

\usepackage{graphics} 
\usepackage{epsfig} 
\usepackage{mathptmx} 
\usepackage{times} 
\usepackage{amsmath} 
\usepackage{amssymb}  

\usepackage{booktabs}
\usepackage{xcolor}
\usepackage{tabularx}

\usepackage{censor}
\usepackage{subcaption}
\usepackage{hyperref}
\hypersetup{
    colorlinks=true,
    linkcolor=blue,
    citecolor=blue,
    urlcolor=blue
}
\usepackage{svg} 
\usepackage{multirow} 
\usepackage{balance}
\usepackage{algorithm}
\usepackage{algpseudocode}
\usepackage{threeparttable}
\usepackage{tikz, graphicx}
\usetikzlibrary{arrows.meta, positioning, calc}

\title{\LARGE \bf
Real-Time Diffusion Relighting as a Neural Sensor: \\ Closing the Low-Light Domain Gap for Robotic Perception
}

\StopCensoring

\author{
\censor{Shengjie Xu}\textsuperscript{1}\thanks{
  \textsuperscript{1}\censor{University of} \censor{Maryland,} \censor{College Park,} 
  \texttt{\censor{\{sjxu,} \censor{awgao,} \censor{lin\}@umd.edu}}. 
  \textsuperscript{2}\censor{Dolby Laboratories,} 
  \texttt{\censor{\{anshul.rai,} \censor{daeyeol.lee,} \censor{guanming.su\}@dolby.com}}.}, 
\censor{Alexander Gao}\textsuperscript{1}, 
\censor{Anshul Rai}\textsuperscript{2}, 
\censor{Dae Yeol Lee}\textsuperscript{2}, 
\censor{Guan-Ming Su}\textsuperscript{2}, 
\censor{Ming C. Lin}\textsuperscript{1}
}

\begin{document}

\maketitle
\thispagestyle{empty}
\pagestyle{empty}

\begin{abstract}

Low-light conditions severely degrade robotic perception: object detectors trained on daylight imagery are prone to failure under dark and adverse weather conditions that fall outside of their training distribution. Furthermore, from an efficiency standpoint, existing Low-Light Image Enhancement (LLIE) methods are designed for offline processing and are therefore impractical to integrate into live control loops. To address these issues, we propose to embed a {\em Generative Neural Sensor} within the control loop that treats illumination as a generative prior rather than a signal processing problem, enabling task-relevant semantic recovery. We further propose an \emph{Real-Time Perceptual Architecture} that decouples high-latency semantic recovery from low-latency robotic control. Rather than blocking the control loop, our system runs a large-scale image-to-image diffusion model in a background thread, and is activated only when a drop in detection confidence signals a \emph{perceptual uncertainty event}, significantly reducing computational overhead. Our method sets a new state-of-the-art performance on the ExDark~\cite{loh2019exdark} low-light benchmark with 87.1\% mAP, and under adverse weather conditions, significantly boosts detection rates over competing methods. On a navigation task in a closed-loop autonomous driving simulator (CARLA), we achieve a 100\% lap completion rate, establishing a simulation-based proof-of-concept for integrating high-latency generative models with standard robotic control autonomy stacks.
\end{abstract}


\section{INTRODUCTION}

Autonomous robots operating in low-light environments face severe perception challenges, as low signal-to-noise ratios and harsh shadows from active illumination confuse downstream object detectors~\cite{darkgs2024}. To address this, low-light image enhancement (LLIE) methods typically train feed-forward networks to minimize pixel-wise reconstruction error (e.g., PSNR, SSIM) against well-lit ground truth~\cite{liu2025mambasflnet}. However, this signal-processing approach often aggressively denoises the image, destroying the faint, residual contrast edges that downstream perceptual algorithms rely on~\cite{liu2020connecting}. Furthermore, state-of-the-art LLIE methods remain too computationally demanding for real-time operation and are designed as offline filters rather than in-the-loop robotic sensors.

We posit a fundamental shift: rather than treating illumination removal as a signal-processing problem, we formulate it as a generative task. Our core insight is that large-scale Latent Diffusion Models (LDMs) encode a rich \textit{semantic} lighting prior from the scale of their training distribution. When appropriately adapted, these models do not merely adjust pixel intensities; they reconstruct plausible, task-relevant object geometries and structural layouts occluded by darkness or severe degradation, shifting the objective from human-centric pixel fidelity to robot-centric semantic recovery.

To harness this heavy computational prior for live robotic operation, we deploy generative relighting as a \textit{neural sensor} within an asynchronous, active perception architecture. Motivated by Bajcsy's active perception paradigm~\cite{bajcsy2018active}, our framework treats the high-latency generative model not as a continuous feed-forward pipeline, but as an intentional computational ``action'' taken only to resolve genuine semantic ambiguity. By monitoring the robot's baseline detection confidence, our system identifies Perceptual Uncertainty Events, triggering the generative sensor in an asynchronous background thread only when the primary perception stack fails, decoupling enhancement latency from the vehicle's high-frequency control loop.
%
%
%
%

\begin{figure}[t]
\centering
\resizebox{\columnwidth}{!}{%
\begin{tikzpicture}[
    >=Stealth,
    imgnode/.style={inner sep=0pt, outer sep=0pt},
    methodpill/.style={
        draw=black!50, fill=blue!5, rounded corners=4pt,
        font=\scriptsize\bfseries, align=center,
        inner xsep=5pt, inner ysep=3pt
    },
    loratag/.style={
        draw=orange!70!black, fill=orange!10, rounded corners=3pt,
        font=\tiny\bfseries, inner xsep=4pt, inner ysep=2pt
    },
    fpstag/.style={
        draw=teal!60!black, fill=teal!8, rounded corners=6pt,
        font=\tiny, inner xsep=5pt, inner ysep=1.5pt
    },
    restag/.style={
        rounded corners=3pt, font=\tiny, align=center,
        inner sep=3pt, minimum width=2.8cm
    },
    sublabel/.style={font=\tiny, text=black!55},
]

\node[imgnode] (darkimg) at (0, 0) {%
    \includegraphics[width=2.cm, height=2.cm]{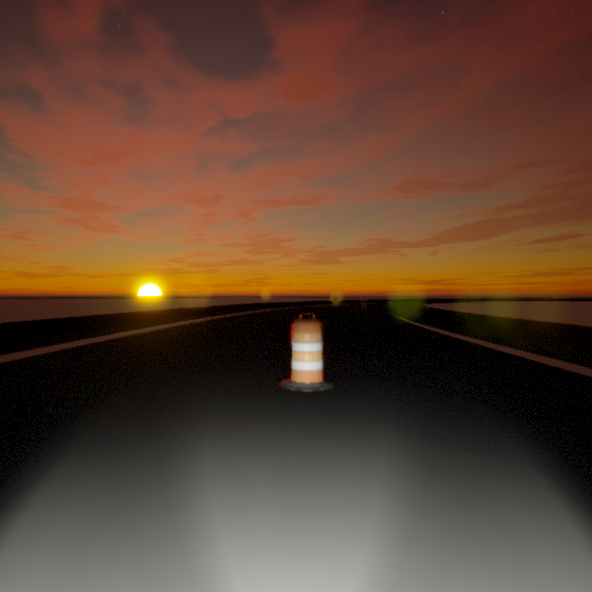}
};

\node[imgnode, right=2.6cm of darkimg] (relitimg) {%
    \includegraphics[width=2.cm, height=2.cm]{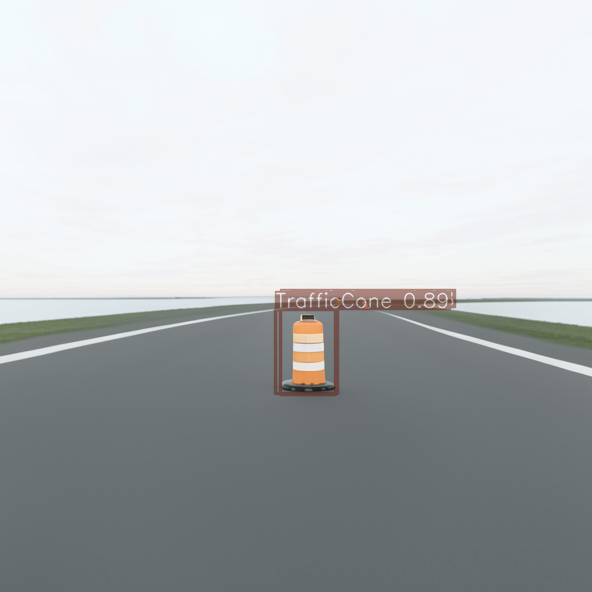}
};



\node[below=3pt of darkimg, font=\tiny\bfseries] (dlabel) {Dark input};
\node[below=3pt of relitimg, font=\tiny\bfseries] (rlabel) {Relit output};

\coordinate (midpoint) at ($(darkimg.east)!0.5!(relitimg.west)$);

\node[methodpill] (method) at (midpoint) {Neural\\Sensor};
\node[loratag, below=2pt of method] (lora) {LoRA};

\draw[->, line width=0.8pt, black!55]
    ([xshift=2pt]darkimg.east) -- ([xshift=-2pt]method.west);
\draw[->, line width=0.8pt, black!55]
    ([xshift=2pt]method.east) -- ([xshift=-2pt]relitimg.west);


    
\node[fpstag, below=2pt of lora] (fps)
    {16\,FPS $\cdot$ 62\,ms};






\end{tikzpicture}%
}
\caption{%
Low-light conditions cause detection failures in autonomous 
navigation (left). Our \textit{Generative Neural Sensor} 
recovers daylight-quality semantics on-demand (right) via an 
asynchronous diffusion prior, restoring obstacle detection 
at 62\,ms without blocking the control loop.
}

\label{fig:teaser}
\end{figure}

The main contributions of this paper are:
\begin{itemize}
    \item A {\em generative neural sensing} module that recovers task-relevant scene semantics and object structure under adverse illumination, leveraging large-scale diffusion priors.
    \item An {\em uncertainty-aware activation mechanism} and {\em asynchronous architecture} that treats generative enhancement as a sparse computational resource, dynamically allocating compute to reduce average overhead by approximately 80\%.
    \item Empirical evidence demonstrating that generative semantic priors outperform traditional pixel-reconstruction LLIE objectives for downstream detection tasks under genuine perceptual occlusion.
    \item Open-source, closed-loop validation within an autonomous driving simulator, demonstrating feasibility for future deployment on physical robotic hardware.
\end{itemize}


%
%

\section{RELATED WORK}

\begin{figure*}[t]
    \centering
    \includegraphics[width=0.6\textwidth]{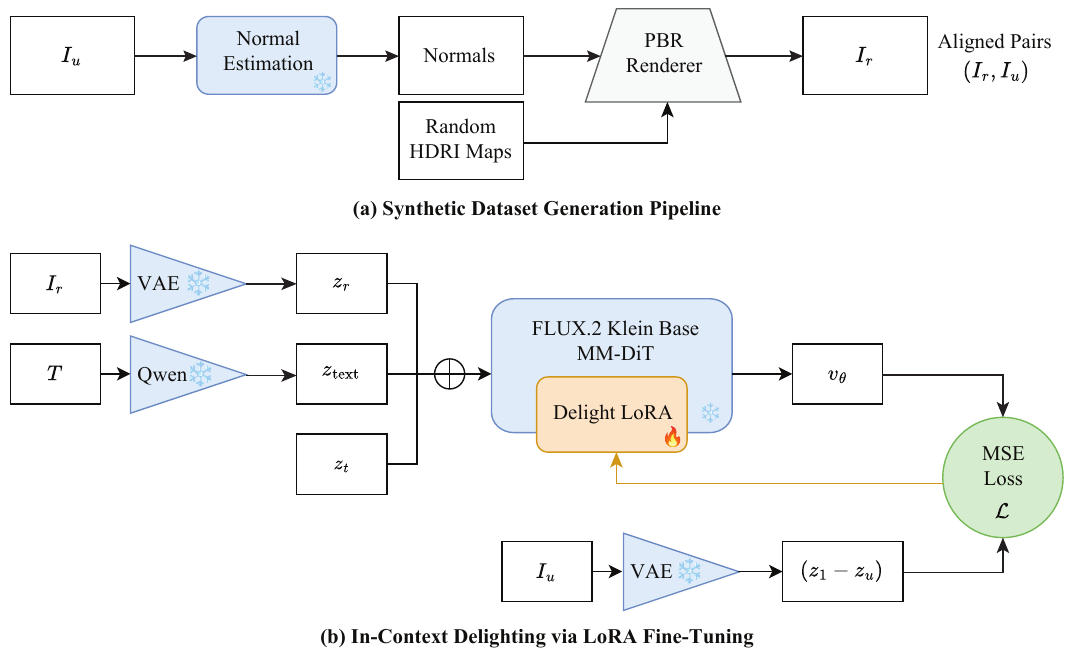}
    \caption{Overview of the Generative Neural Sensor Pipeline. (a) Synthetic Data Engine: Extracting geometry to render pixel-aligned training pairs ($I_u$, $I_r$). (b) Asynchronous Active Perception: triggered by perceptual uncertainty events, the generative backbone is specialized for illumination removal using task-specific adapters.}
    \label{fig:architecture}
\end{figure*}

\subsection{Low-Light Image Enhancement}
Traditional methods for low-light enhancement rely on histogram equalization, gamma correction, or Retinex theory to adjust pixel intensities. While computationally inexpensive, these amplify noise and fail to recover semantic information in pitch-black regions \cite{guo2020zero}. Deep learning approaches such as Zero-DCE \cite{guo2020zero} and EnlightenGAN \cite{jiang2021enlightengan} improve visual quality but often introduce temporal artifacts that degrade downstream perception. DarkGS \cite{darkgs2024} applies 3D Gaussian Splatting to dark scenes but requires offline multi-view optimization, making it unsuitable for single-shot online navigation. More recently, MambaSFLNet \cite{liu2025mambasflnet} proposes a selective frequency learning network using Mamba state-space models for joint low-light enhancement and object detection, achieving 25.64~Hz throughput. While such specialized architectures are performant, their deterministic nature fundamentally relies on existing pixel data, causing them to struggle when object geometries are heavily occluded by severe degradations (e.g.\ dense fog). Our approach instead harnesses the generative semantic priors of a foundational Latent Diffusion Model to reconstruct task-relevant layouts. Since large-scale generative models are inherently more computationally demanding, we deploy the model within an asynchronous, event-driven architecture, acting as an on-demand neural sensor, that decouples its execution latency from the primary control loop.

\subsection{Generative AI for Robotic Perception}
The direct application of large-scale Diffusion Transformers for \textit{per-frame} robotic perception remains largely unexplored due to prohibitive inference latency. Diffusion Policies \cite{chi2023diffusion} and DiffuseLoco \cite{zhuang2024diffuseloco} utilize generative models for action and trajectory planning from sensory input, but act as policy controllers rather than perceptual sensors. Generative Grasping CNNs \cite{morrison2018closing} demonstrate real-time closed-loop control at up to 50~Hz but target specialized manipulation tasks using small-scale, task-specific networks. Similarly, prior work utilizing diffusion models for offline video prediction or rendering (e.g.\ Cosmos-Predict \cite{nvidia2026worldsimulationvideofoundation}) acts as an offline simulator. Crucially, no prior work deploys a large-scale diffusion model \textit{inside} a live control loop as an asynchronous active perception sensor for environmental condition recovery. Our work addresses this gap by placing a generative relighting model directly within a robotic navigation stack, treating generative enhancement as a sparse, on-demand resource invoked only when detection confidence signals perceptual uncertainty.

\subsection{Diffusion-Based Relighting}
Recent diffusion-based relighting methods have demonstrated strong illumination control for image synthesis. SynthLight \cite{hou2024synthlight} frames the task as high-fidelity re-rendering in response to modeled environmental lighting changes. DiLightNet \cite{zeng2024dilightnet} introduces a three-stage pipeline for fine-grained lighting control using radiance hints derived from depth. IC-Light \cite{zhangscaling} tackles illumination harmonization with a diffusion model trained for in-the-wild consistency, while Neural Gaffer \cite{jin2024neural} and ScribbleLight \cite{choi2024scribblelightsingleimageindoor} extend single-image relighting to category-agnostic and user-guided settings. However, all of these methods target static aesthetic editing or offline rendering, lacking the mechanisms for low-latency, task-driven illumination recovery required by autonomous systems. Ours is the first to deploy a diffusion-based relighting model as a neural sensor within a live robotic navigation stack.

\subsection{Active Perception and Event-Driven Architectures}
Active perception dictates that a robot should take physical actions to maximize information gain and minimize uncertainty \cite{bajcsy2018active}. We extend this to the computational domain: the ``action'' is the state-dependent allocation of computational resources---specifically, deploying a heavy generative prior to resolve perceptual ambiguity. Adaptive test-time compute scaling has demonstrated outsized performance gains in foundation models \cite{snell2024scaling}, yet remains largely confined to textual reasoning. We bridge this to robotics via an event-driven perceptual architecture. While recent works have applied event-driven lighting to guide attention in immersive virtual environments \cite{10972705}, and others use language models to detect narrative events \cite{gao2023exploring}, we adapt this paradigm for machine perception. Specifically, we utilize YOLO detection confidence as an empirical proxy for perceptual reliability \cite{kendall2017uncertainties}, triggering a generative relighting ``event'' to restore lost semantic information. By activating the high-latency neural sensor \textit{only} when confidence drops below a specified threshold, our system treats generative enhancement as a sparse computational resource, decoupling the perception overhead from the vehicle's control frequency.

\section{METHODOLOGY}

\subsection{Formalism: Generative Neural Sensor Architecture}

We propose to elevate the generative prior of a large-scale rectified flow transformer as a \textit{neural sensor} for robotic perception. The architecture dictates a four-component pipeline---Input, Trigger, Processor, and Consumer---that decouples high-latency generative enhancement from the low-latency robot control loop.

\subsubsection{Rectified Flow Matching}
Diffusion-based generative models traditionally rely on stochastic differential equations to transport Gaussian noise to data~\cite{ho2020denoising,song2020score}. Flow Matching~\cite{liu2022flow} replaces this with a deterministic linear interpolation between noise $x_1$ and the clean image $x_0$ at time $t \in [0,1]$:
\begin{equation}
    x_t = t\, x_1 + (1-t)\, x_0
\end{equation}
A neural network $v_\theta$ predicts the velocity field along this straight-line trajectory by minimizing:
\begin{equation}
    \mathcal{L}_{\text{flow}}(\theta) = \mathbb{E}_{t, x_0, x_1} \left[ \| v_\theta(x_t, t, c) - (x_1 - x_0) \|^2 \right]
\end{equation}
Straight-line trajectories efficiently integrate in fewer steps than curved diffusion paths.

\subsubsection{In-Context Illumination Recovery}
Image enhancement is formulated as an \textit{in-context generation} task. Rather than using auxiliary conditioning networks~\cite{zhang2023addingconditionalcontroltexttoimage}, a multimodal diffusion transformer jointly processes a reference image and a generation target by concatenating their token sequences~\cite{flux1kontext2025}. Given a conditioning uniformly-lit latent $z_{\text{cond}}$, a noisy target $z_t$, and text embeddings $z_{\text{text}}$, the unified sequence is:
\begin{equation}
    z_{\text{in}} = [z_{\text{text}};\; z_t;\; z_{\text{cond}}]
\end{equation}
The model copies structural content from the reference while transforming it according to the textual generation instruction, preserving structural fidelity.

\subsubsection{Parameter-Efficient Illumination Adapter}
Full fine-tuning of large models risks catastrophic forgetting of pre-trained priors. Low-Rank Adaptation (LoRA)~\cite{hu2021lora} constrains weight updates. For a frozen matrix $W_0 \in \mathbb{R}^{d \times k}$:
\begin{equation}
    W = W_0 + \frac{\alpha}{r} BA
\end{equation}
where $B \in \mathbb{R}^{d \times r}$ and $A \in \mathbb{R}^{r \times k}$ are trainable parameters of rank $r \ll d$. The loss optimizes the adapter to map degraded illumination latents $z_r$ to clean uniformly-lit latents $z_u$:
\begin{equation}
    \mathcal{L}_{\text{delight}} = \mathbb{E}_{t, z_u, z_r, z_1} \left[ \left\| v_\theta([z_t;\, z_r], t, T) - (z_1 - z_u) \right\|^2 \right]
\end{equation}

\subsubsection{Asynchronous Active Perception Pipeline}
\label{sec:event_driven}

Grounded in active perception~\cite{bajcsy2018active}, the system intentionally deploys the generative model exclusively to resolve semantic ambiguity. The \textbf{Input} is the raw camera frame. The \textbf{Trigger} identifies a \textbf{Perceptual Uncertainty Event} by monitoring the maximum object detection confidence $c_{\max}(t)$. It activates the generative sensor under high uncertainty:
\begin{equation}
    \text{relight}(t) = \begin{cases}
      1 & \text{if } c_{\max}(t) < \tau_{\text{low}} \\
      0 & \text{otherwise}
   \end{cases}
\end{equation}
The \textbf{Processor} runs the generative backbone and illumination adapter in an asynchronous background thread, non-blocking to the main loop. The \textbf{Consumer} feeds the recovered semantic image back to the real-time detector and the downstream collision avoidance controller. This maps generative perception to a sparse, on-demand computational architecture using hysteresis thresholding (where deactivation occurs when $c_{\max} > \tau_{\text{high}}$) to prevent output oscillation.

\subsection{Implementation}
\label{sec:implementation}

We instantiate the framework using specific foundation models and robotic simulators. 

\subsubsection{Generative Backbone and Adapter}
To satisfy the stringent latency constraints of real-time robotic deployment, we adopt the highly efficient FLUX.2-klein-4B~\cite{flux2klein2026, flux2klein4b_repo} as our frozen large-scale rectified flow transformer, which natively supports sequence concatenation. We augment this backbone with a parameter-efficient Low-Rank Adapter (LoRA)~\cite{hu2021lora} ($r{=}16$, $\alpha{=}4$, 0.8\% trainable weights). Our training protocol adapts the framework of Tsaban~\cite{tsaban2025flux2kleinrelight}, employing the DreamBooth architecture~\cite{ruiz2023dreambooth} in conjunction with LoRA. While we utilize the identical synthetic \texttt{flux-delight-dataset}---generated via normal estimation and physically based HDRI rendering (see Fig.~\ref{fig:architecture}a)---our implementation is independently developed using the Diffusers library~\cite{von-platen-etal-2022-diffusers}. Crucially, to enable in-the-loop deployment, our approach diverges from prior offline methods by utilizing the smaller 4B-parameter base model and employing fp8 mixed-precision training to fundamentally minimize computational overhead. The adapters are optimized using AdamW ($\beta_1{=}0.9$, $\beta_2{=}0.999$, lr=$1{\times}10^{-4}$) for 1,280 steps on a single NVIDIA RTX PRO 6000 Blackwell GPU.

The flow is steered by a natural language relighting instruction $T$\footnote{The exact prompt utilized is: \textit{``Relight the image to remove all existing lighting and replace with neutral, uniform illumination.''}}. By isolating gradient updates to the adapters, the model acts as a learned inverse rendering operator that robustly transfers task-relevant structural layouts.

\subsubsection{System Deployment}
The \textbf{Trigger} uses a confidence threshold of $\tau_{\text{low}} = 0.3$ with a hysteresis deactivation of $\tau_{\text{high}} = 0.5$. The \textbf{Consumer} employs a frozen YOLO26 detector~\cite{yolo26_ultralytics} as the real-time object detector. The downstream vehicle closed-loop autonomous navigation stack is implemented in CARLA~\cite{dosovitskiy2017carla} on the Thunderhill Raceway environment~\cite{thunderhill2025}, using an MPC\_DLC formulation capable of adjusting its lateral offset based on the semantically recovered detections. The processor background thread comfortably achieves 16 FPS throughput under full system integration.

\begin{figure*}[!t]
    \centering
    \setlength{\tabcolsep}{1pt} 
    \renewcommand{\arraystretch}{0.2} 

    \begin{tabular}{cccccc} 
        \rotatebox{90}{\hspace{1.8em}\scriptsize\textbf{Original}} &
        \subfloat{\includegraphics[width=0.18\linewidth]{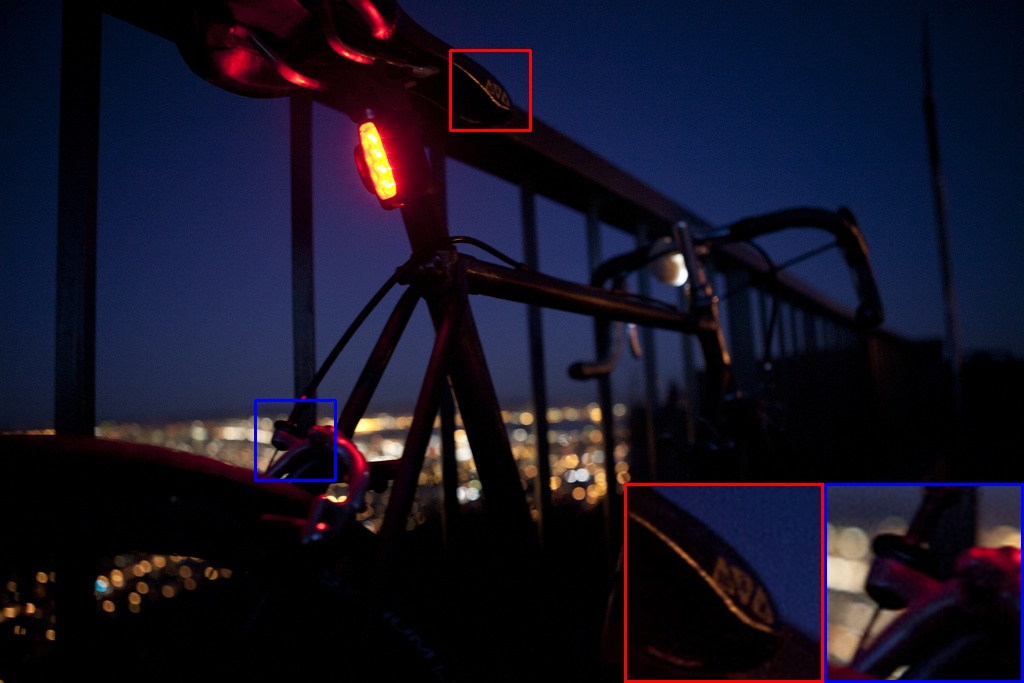}} &
        \subfloat{\includegraphics[width=0.18\linewidth]{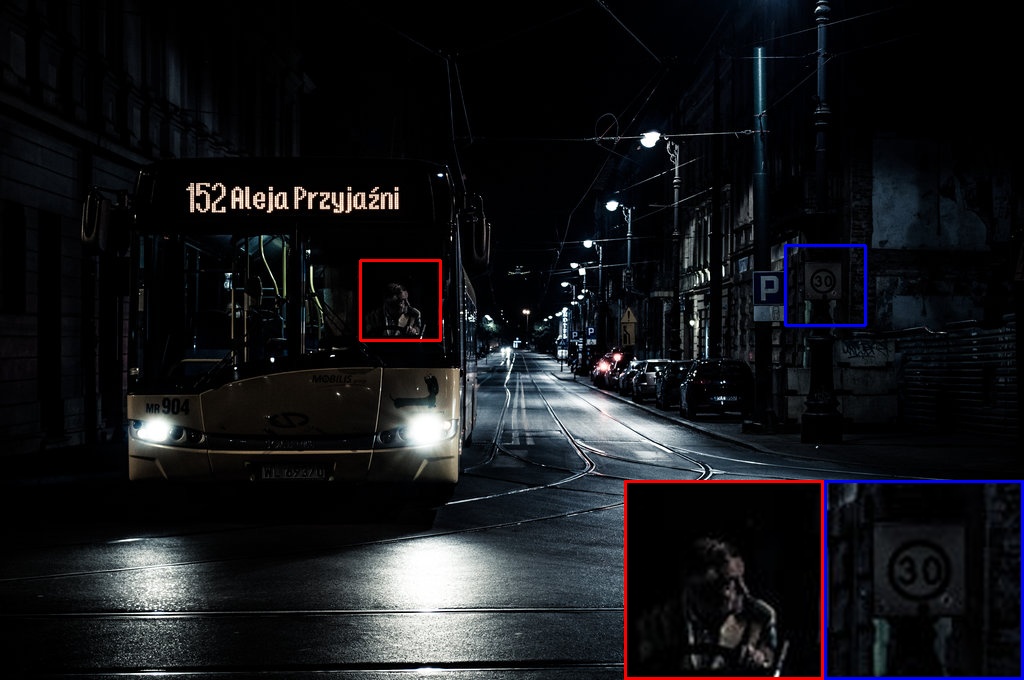}} &
        \subfloat{\includegraphics[width=0.18\linewidth]{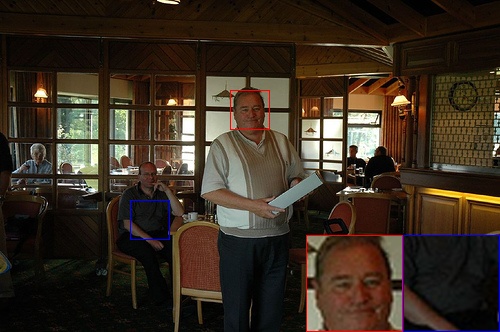}} &
        \subfloat{\includegraphics[width=0.18\linewidth]{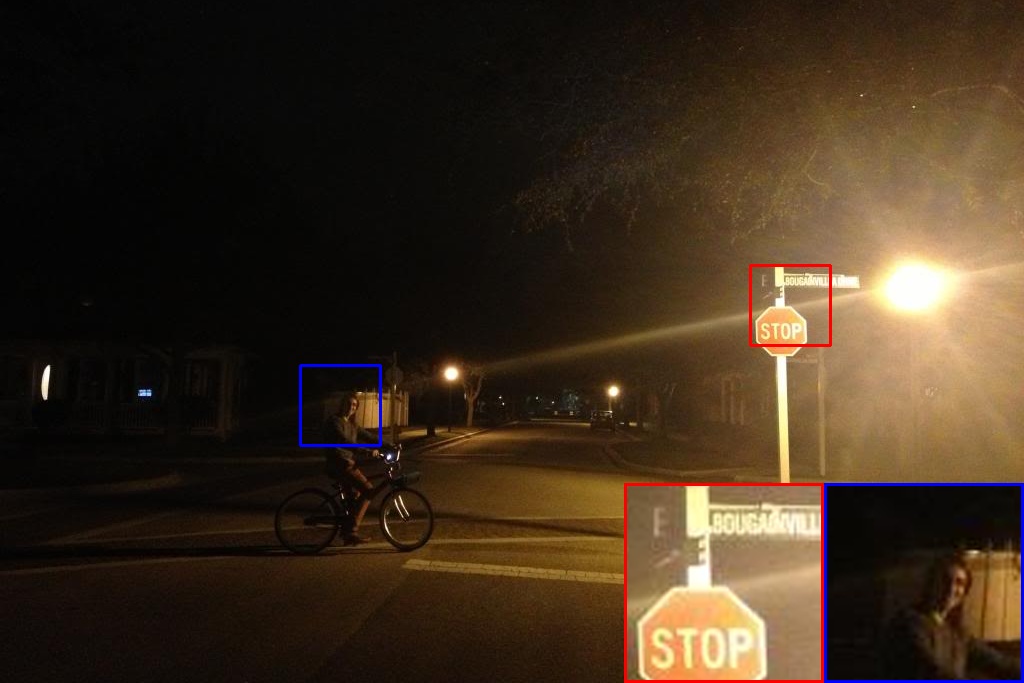}} &
        \subfloat{\includegraphics[width=0.18\linewidth]{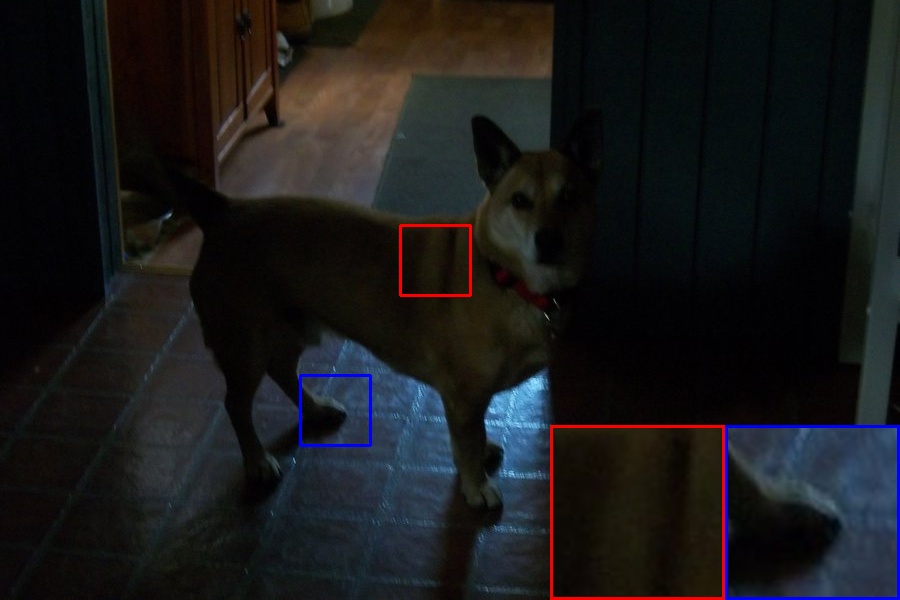}} \\[-0.0ex]

        \rotatebox{90}{\hspace{1.8em}\scriptsize\textbf{Mamba}} &
        \subfloat{\includegraphics[width=0.18\linewidth]{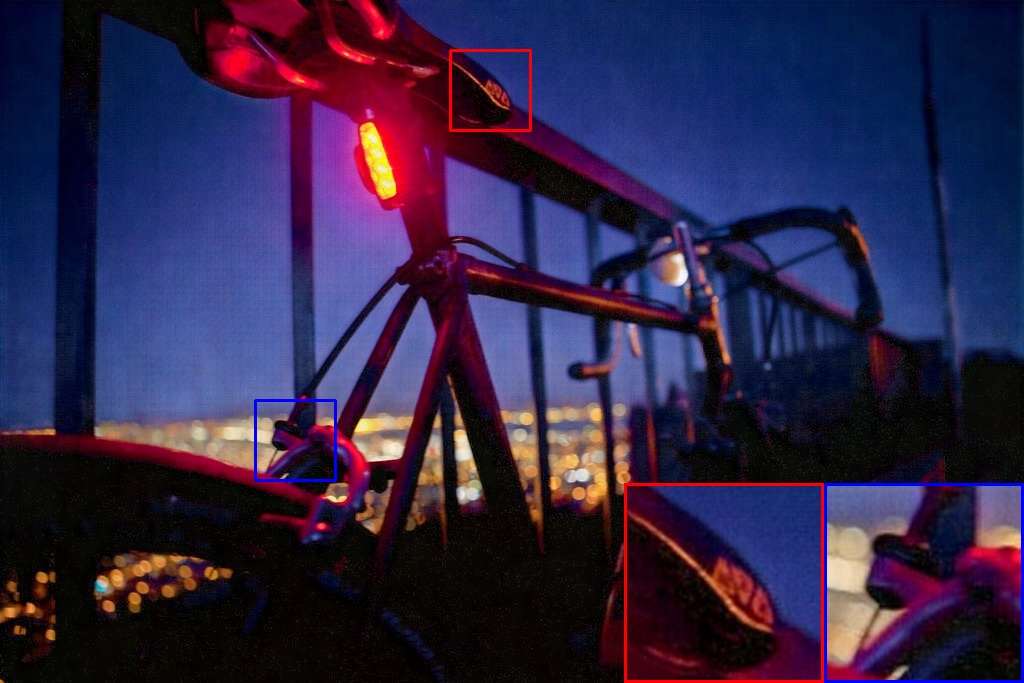}} &
        \subfloat{\includegraphics[width=0.18\linewidth]{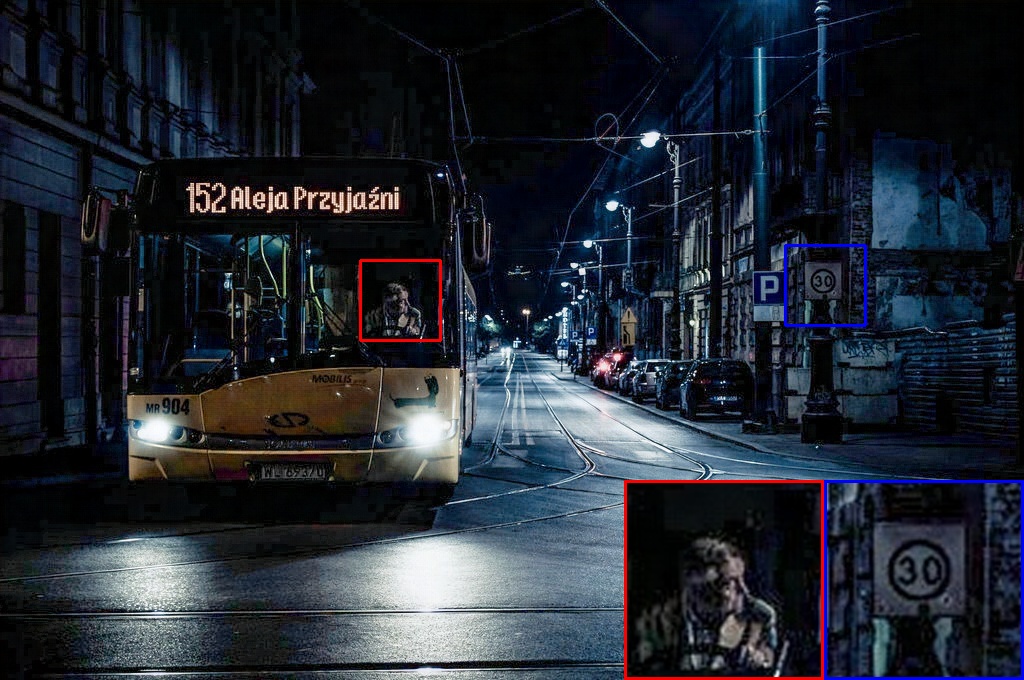}} &
        \subfloat{\includegraphics[width=0.18\linewidth]{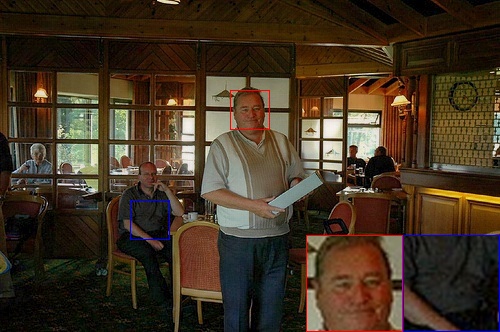}} &
        \subfloat{\includegraphics[width=0.18\linewidth]{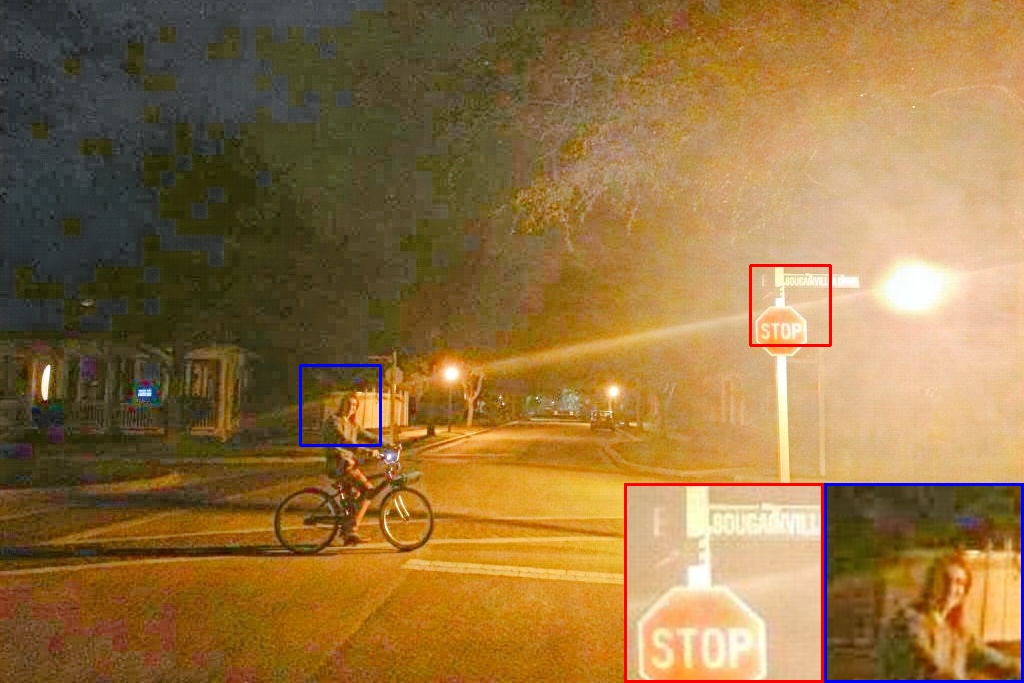}} &
        \subfloat{\includegraphics[width=0.18\linewidth]{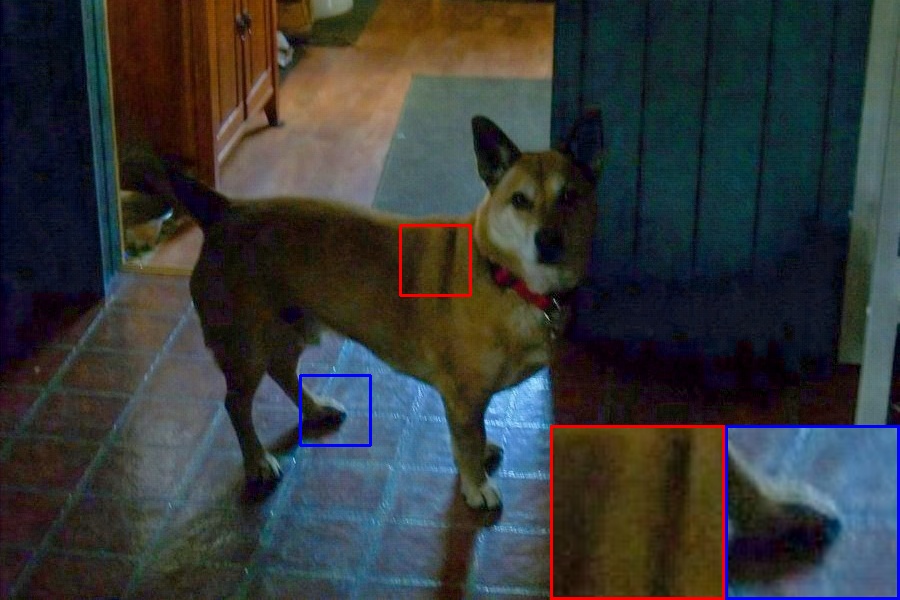}} \\[-0.0ex]

        \rotatebox{90}{\hspace{1.8em}\scriptsize\textbf{Ours}} &
        \subfloat{\includegraphics[width=0.18\linewidth]{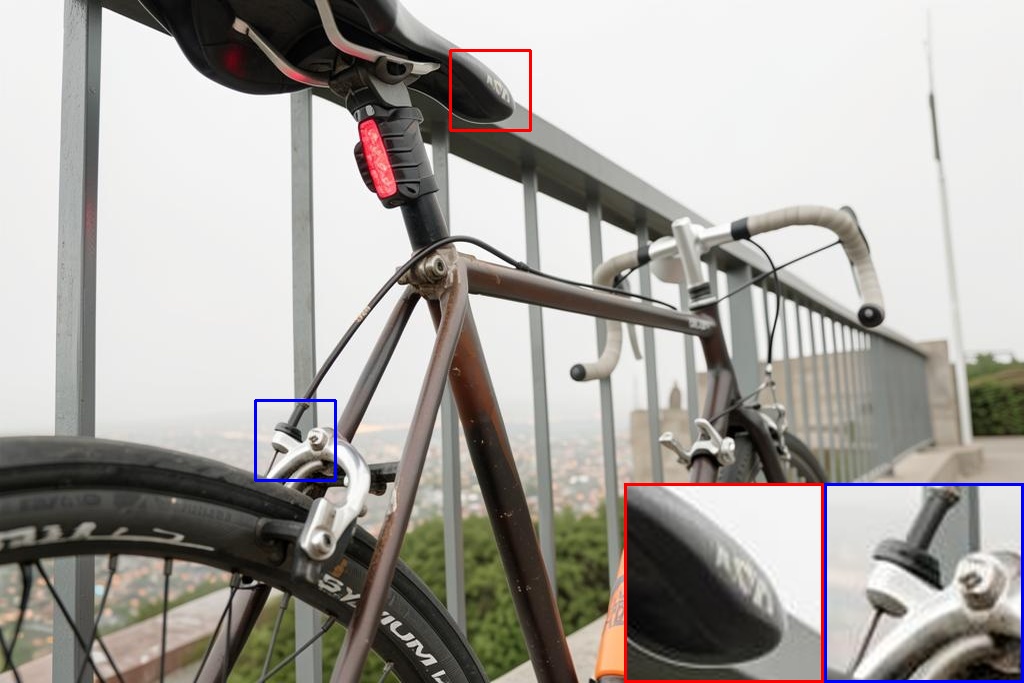}} &
        \subfloat{\includegraphics[width=0.18\linewidth]{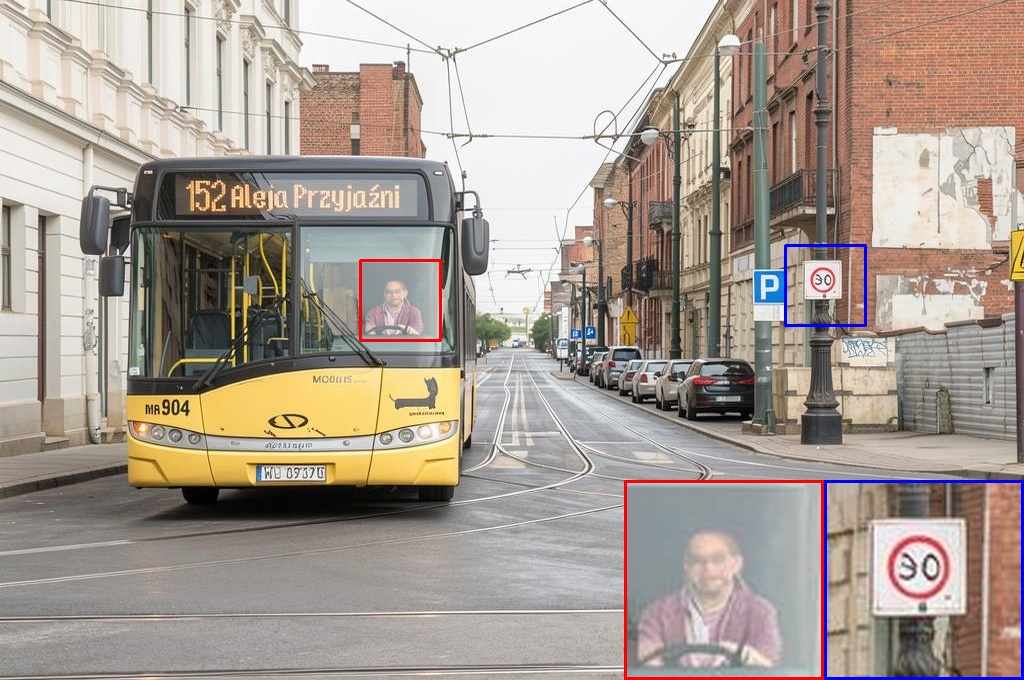}} &
        \subfloat{\includegraphics[width=0.18\linewidth]{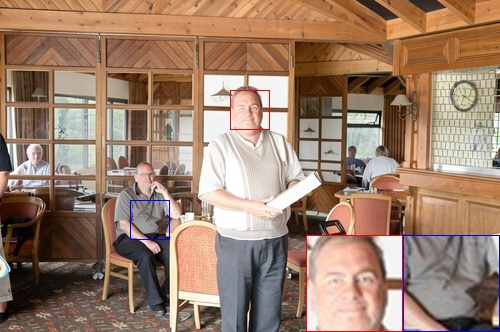}} &
        \subfloat{\includegraphics[width=0.18\linewidth]{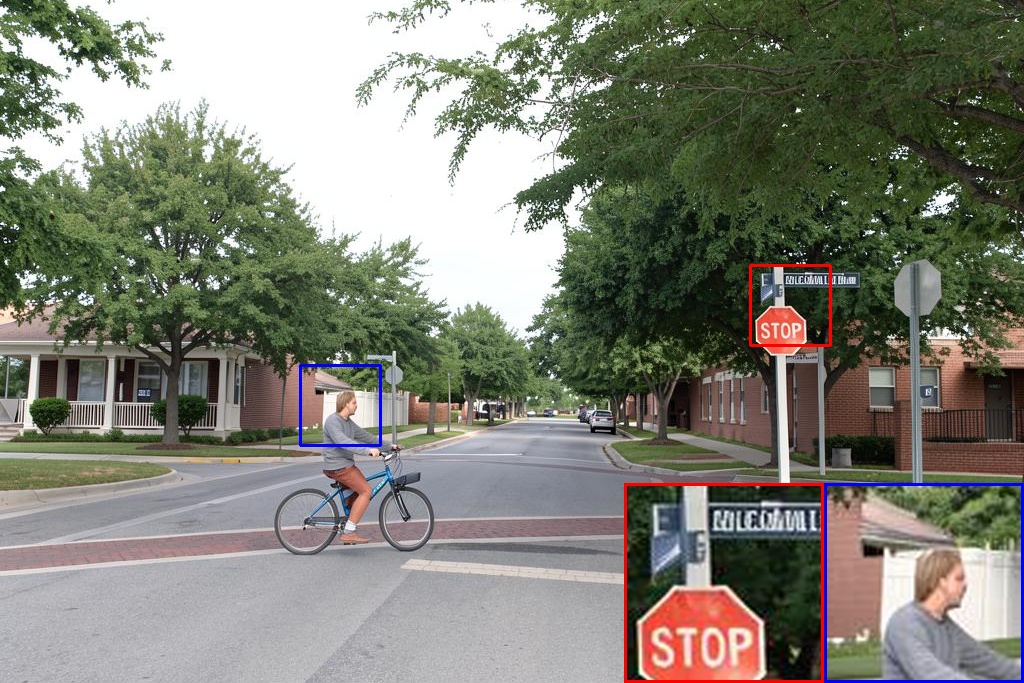}} &
        \subfloat{\includegraphics[width=0.18\linewidth]{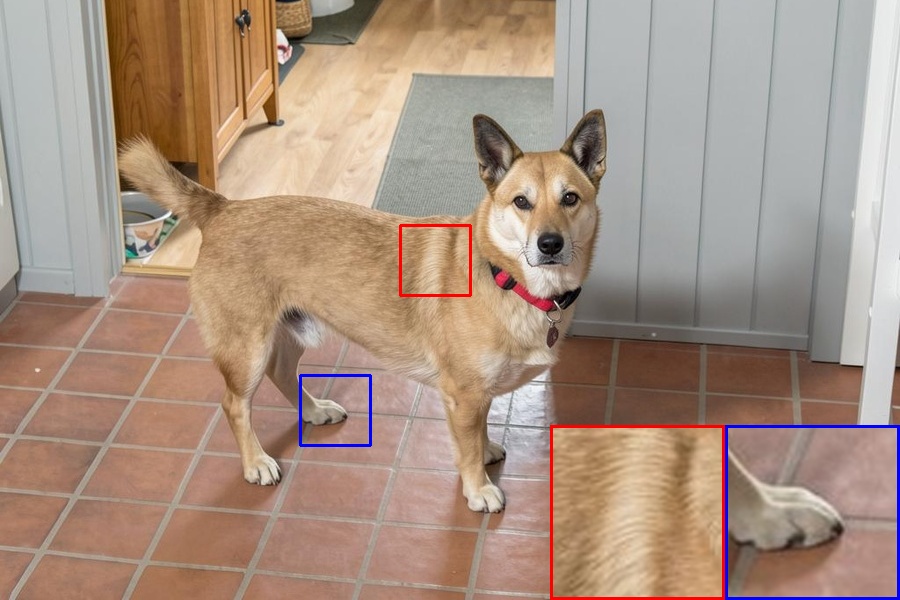}} \\

    \end{tabular}
    \caption{Qualitative comparison of detection performance on ExDark~\cite{loh2019exdark}. From top to bottom: Original low-light input, MambaSFLNet\cite{liu2025mambasflnet}, and ours. Our method successfully recovers semantic layouts and object detections in extreme darkness.}
    \label{fig:comparison_transpose}
    
\end{figure*}

\section{EXPERIMENTS}

Our evaluation is presented in three stages.
\textbf{(1)} We first establish that our relighting pipeline improves detection on real-world low-light imagery, benchmarked against LLIE SOTA on ExDark (§\ref{subsubsec:exdark}).
\textbf{(2)} We then ask whether this benefit transfers to the autonomous driving domain, where a synthetic domain gap exists between the Delight LoRA's training distribution (natural images) and driving scenes; we test this on vKITTI2 (§\ref{subsubsec:vkitti2}). \textbf{(3)} Finally, we close the loop with a live CARLA deployment, validating the full system for real-time robotic operation. To rigorously test our hypothesis, the downstream detector is fine-tuned exclusively on clear daytime simulator data, ensuring it remains strictly dependent on the generative sensor to recover perception in low-light conditions (§\ref{subsec:dynamic_eval}).

\subsection{Static Evaluation}
\label{subsec:static_eval}

Static evaluation isolates the relighting module's effect on object detection quality, independent of control or navigation variables. We apply our relighting pipeline as a fixed preprocessing step and measure YOLO detection performance on two complementary benchmarks. All methods are evaluated using YOLO26s~\cite{yolo26_ultralytics} with standard detection metrics: {\em Precision, Recall, F1-Score, and mAP@50}. We compare against Raw Baseline (no enhancement), CLAHE~\cite{pizer1990contrast}, MambaSFLNet~\cite{liu2025mambasflnet} (IROS 2025 SOTA), and Ours. 

\subsubsection{ExDark Real-World Validation}
\label{subsubsec:exdark}

We evaluate on ExDark~\cite{loh2019exdark}, a real-world low-light benchmark of 7,363 images spanning 12 object categories captured under diverse natural low-light conditions including noise, blur, and color distortion.

Following MambaSFLNet~\cite{liu2025mambasflnet} and RetinexNet~\cite{wei2018deep}, we adopt the standard 5,890/1,473 train/test split for direct comparison. Our relighting pipeline is applied as a fixed preprocessing module to both splits. Crucially, YOLO-v3~\cite{redmon2018yolov3} is trained from scratch exclusively on the \textit{relit} training images (generatively restored dark images to a daylight distribution) rather than the original dark frames, and evaluated on the equivalently relit test set. This ensures the detector operates strictly within well-lit distributions. Performance is reported as per-class AP and mean AP (mAP@50) across all 12 categories.

\subsubsection{vKITTI2 Domain Gap Evaluation}
\label{subsubsec:vkitti2}
Virtual KITTI 2 (vKITTI2)~\cite{cabon2020vkitti2} provides 
photorealistic synthetic driving scenes with ground-truth 
annotations under five degradation conditions: \textit{Fog}, 
\textit{Rain}, \textit{Sunset}, and \textit{Overcast}. We 
train YOLO26s~\cite{yolo26_ultralytics} exclusively on the 
Clone (clean daylight) split and evaluate on all degraded 
conditions without retraining---simulating a detector trained 
on ideal conditions and deployed into the field. All enhancement 
methods use publicly available pretrained weights without 
domain-specific fine-tuning, ensuring a fair zero-shot 
generalization comparison. The Delight LoRA used in our method 
was trained on natural images rather than driving footage, so 
this benchmark tests whether the detection benefit survives 
the distribution shift to the autonomous driving domain.

\begin{table*}[!t]
\centering
\caption{Generative neural sensors (Ours) outperform specialized low-light enhancement networks across all categories. Low-light detection mAP on ExDark~\cite{loh2019exdark}. \textcolor{red}{Red}: best, \textcolor{blue}{blue}: second best.}
\label{tab:exdark}
\setlength{\tabcolsep}{4pt}
\begin{tabular}{lcccccccccccccc}
\toprule
Methods & Bicycle & Boat & Bottle & Bus & Car & Cat & Chair & Cup & Dog & Motor & People & Table & Mean \\
\midrule
RetinexNet \cite{wei2018deep}      & 0.790 & 0.741 & 0.743 & 0.908 & 0.820 & 0.665 & 0.651 & 0.750 & 0.721 & 0.703 & 0.784 & 0.556 & 0.736 \\
KinD \cite{zhang2019kindling}                  & 0.800 & 0.721 & \textcolor{blue}{0.788} & \textcolor{blue}{0.919} & 0.822 & \textcolor{blue}{0.718} & 0.672 & 0.771 & 0.775 & 0.736 & 0.803 & 0.555 & 0.757 \\
Zero-DCE \cite{guo2020zero}           & 0.806 & 0.750 & 0.762 & 0.914 & \textcolor{blue}{0.837} & 0.681 & 0.677 & 0.769 & \textcolor{blue}{0.788} & 0.728 & 0.801 & 0.535 & 0.754 \\
Retinexformer \cite{cai2023retinexformer}& 0.809 & 0.769 & 0.753 & 0.914 & 0.814 & 0.688 & 0.689 & 0.763 & 0.766 & 0.769 & 0.805 & 0.543 & 0.757 \\
MambaIR \cite{guo2024mambair}            & 0.803 & 0.763 & 0.752 & 0.903 & 0.830 & 0.687 & 0.684 & 0.761 & 0.721 & 0.738 & \textcolor{blue}{0.813} & 0.556 & 0.751 \\
MambaSFLNet \cite{liu2025mambasflnet}   & \textcolor{blue}{0.821} & \textcolor{blue}{0.775} & 0.773 & 0.891 & 0.824 & 0.665 & \textcolor{blue}{0.696} & \textcolor{blue}{0.820} & 0.779 & \textcolor{blue}{0.783} & 0.809 & \textcolor{blue}{0.558} & \textcolor{blue}{0.766} \\
\midrule
Ours & \textcolor{red}{0.886} & \textcolor{red}{0.883} & \textcolor{red}{0.867} & \textcolor{red}{0.989} & \textcolor{red}{0.878} & \textcolor{red}{0.895} & \textcolor{red}{0.854} & \textcolor{red}{0.863} & \textcolor{red}{0.933} & \textcolor{red}{0.827} & \textcolor{red}{0.851} & \textcolor{red}{0.729} & \textcolor{red}{0.871} \\
\bottomrule
\end{tabular}
\vspace*{-1em}
\end{table*}

\subsection{Dynamic Evaluation: Real-Time Deployment in CARLA}
\label{subsec:dynamic_eval}

The dynamic evaluation validates that our event-driven generative pipeline enables safe, real-time autonomous navigation under severe lighting degradation. To isolate the contribution of the relighting module, the downstream YOLO26 detector~\cite{yolo26_ultralytics} is fine-tuned exclusively on clear daylight simulator data. Consequently, the detector depends entirely on our generative sensor to recover operational semantics when deployed in low-light environments. 

We deploy the closed-loop system in the CARLA simulator~\cite{dosovitskiy2017carla} on the Thunderhill Raceway track~\cite{Kapania_2016,thunderhill2025}, utilizing a forward-facing camera rig. Static traffic cones are stochastically seeded along the reference trajectory at a rate of $\rho = 0.02$ per waypoint, yielding approximately 90 obstacles per lap on average. The autonomous agent governs its longitudinal velocity via a standard PID controller and executes lateral evasive maneuvers using our predictive MPC\_DLC formulation.
 A deterministic finite state machine
handles out-of-bounds fault recovery to enable continuous multi-lap benchmarking on an NVIDIA RTX Pro 6000 Blackwell GPU. 

To ensure rigorous benchmarking across lighting conditions, we define three standardized time-of-day environments for our testing: Day ($\circ$, 12:00), Sunset ($\oslash$, 18:00), and Night ($\bullet$, 22:00). For each experimental condition, performance metrics represent the average and variance calculated over five complete laps. Crucially, we utilize fixed random seeds to guarantee identical stochastic obstacle placement across all lighting setups, proving that any performance variation stems strictly from perceptual degradation rather than structural layout differences. We structure our evaluation into three targeted experiments:

\subsubsection{Cross-Controller Navigation }
To validate that the generative sensor's semantic outputs successfully drive downstream physical reactions, we compare two active controllers (Baseline MPC vs. MPC\_DLC). Both are evaluated under Night ($\bullet$) conditions with continuous generative relighting engaged. This test isolates whether the dynamic lateral offset successfully leverages the relit perception manifold to physically avoid detected obstacles compared to naive static tracking.

\subsubsection{Closed-Loop Navigation Safety}
Our capstone evaluation proves the necessity of the generative neural sensor under severe degradation. Using the optimized MPC\_DLC controller, we benchmark identical continuous 5-lap runs across perfect lighting equivalent to a Day ($\circ$) Oracle, intermediate degradation at Sunset ($\oslash$), and absolute failure at Night ($\bullet$). We measure the obstacle collision rate across the unenhanced baseline, standard mathematical contrast stretching (CLAHE), and our generative sensor. The goal is to prove whether the diffusion prior can fully restore the daytime safety margin when standard sensing fails.

\subsubsection{Event-Driven Triggering Efficiency}
To prove that treating generative enhancement as a sparse resource drastically reduces computational overhead, we evaluate the event-driven logic under transitional Sunset ($\oslash$) lighting. We implement a hysteresis threshold on the detector's maximum confidence score (activating when $c_{\max} < 0.3$ and deactivating when $c_{\max} > 0.5$). We track the empirical activation rate ($\alpha$) to compute the true average latency reduction without compromising the established navigation safety.

\section{RESULTS}
\label{sec:results}


\subsection{Static: ExDark Real-World Validation}

Table~\ref{tab:exdark} reports per-class AP and mean AP on ExDark, following the train/test split of MambaSFLNet~\cite{liu2025mambasflnet} and RetinexNet~\cite{wei2018deep} for direct comparison. This experiment establishes the core effectiveness of our relighting pipeline on genuine low-light imagery before any domain-specific considerations.

Our method achieves \textbf{87.1\% mean AP}, outperforming MambaSFLNet~\cite{liu2025mambasflnet} by \textbf{10.5\%}. Unlike MambaSFLNet, which leads in only 6 of 12 categories, our approach achieves the highest AP across \textit{all} 12 categories, demonstrating consistent gains. Notable improvements are observed in Bus (0.989), Dog (0.933), and Cat (0.895), where our relighting pipeline effectively recovers fine-grained details under challenging illumination. Even on harder categories such as Table (0.729) and Motorbike (0.827), our method surpasses all competing approaches, confirming that our pipeline generalizes well to real-world low-light imagery. Please check Fig.~\ref{fig:comparison_transpose} for more qualitative results.

\subsection{Static: Domain Gap Evaluation on vKITTI2}
Table~\ref{tab:vkitti2} reports detection quality across all conditions. \texttt{Fog} induces genuine semantic occlusion: the baseline detector's performance collapses to an F1 score of 0.681 and an mAP of 0.544. Standard contrast stretching via CLAHE provides a partial mathematical recovery (F1: 0.781, mAP: 0.635).

\begin{table}[t]
\centering
\caption{Generative semantic priors recover detection under fog where pixel-reconstruction methods fail. Domain Gap Analysis on vKITTI2~\cite{cabon2020vkitti2}.
\textcolor{red}{Red}: best. \textcolor{blue}{Blue}: second best.}
\label{tab:vkitti2}
\setlength{\tabcolsep}{4pt}
\renewcommand{\arraystretch}{1.1}
\begin{tabular}{llcccc}
\toprule
Cond. & Met. & Baseline & CLAHE~\cite{pizer1990contrast} & MambaSFLNet~\cite{liu2025mambasflnet} & Ours \\
\midrule
\multirow{2}{*}{\texttt{Fog}}
  & F1   & 0.681 & \textcolor{blue}{0.781} & 0.579 & \textcolor{red}{0.855} \\
  & mAP  & 0.544 & \textcolor{blue}{0.635} & 0.452 & \textcolor{red}{0.807} \\
\midrule
\multirow{2}{*}{\texttt{Rain}}
  & F1   & \textcolor{blue}{0.904} & \textcolor{red}{0.916} & 0.872 & \textcolor{black}{0.871} \\
  & mAP  & \textcolor{red}{0.816} & \textcolor{blue}{0.816} & 0.813 & 0.812 \\
\midrule
\multirow{2}{*}{\texttt{Ovrcst}}
  & F1   & \textcolor{red}{0.942} & \textcolor{blue}{0.938} & 0.933 & 0.866 \\
  & mAP  & \textcolor{red}{0.907} & \textcolor{blue}{0.906} & \textcolor{blue}{0.906} & 0.812 \\
\midrule
\multirow{2}{*}{\texttt{Sunset}}
  & F1   & \textcolor{red}{0.927} & 0.926 & \textcolor{blue}{0.922} & 0.864 \\
  & mAP  & \textcolor{red}{0.905} & \textcolor{red}{0.905} & 0.904 & 0.810 \\
\bottomrule
\end{tabular}
\end{table}

Notably, MambaSFLNet---which achieves state-of-the-art PSNR/SSIM on standard LLIE benchmarks~\cite{liu2025mambasflnet}---suffers a severe performance degradation under fog, yielding an F1 of 0.579 and mAP of 0.452, falling \emph{below} the unenhanced baseline. This reveals an optimization mismatch: methods trained to minimize pixel reconstruction error may produce perceptually smooth images but can inadvertently suppress the residual low-contrast structural edges that object detectors rely on. In stark contrast, our method leverages a generative semantic prior to reconstruct plausible object structure rather than minimizing pixel error. This achieves an F1 of \textbf{0.855} and mAP of \textbf{0.807}---a massive gain of 0.174 in F1 over the baseline and 0.276 over MambaSFLNet.

On conditions where the baseline is already robust---\texttt{Rain} (F1: 0.904), \texttt{Overcast} (F1: 0.942), \texttt{Sunset} (F1: 0.927)---all enhancement methods provide negligible benefit, and our method slightly underperforms baseline due to generative artifacts on
already-parseable images. This behavior is not a limitation 
but an empirical motivation for the confidence-triggered 
architecture described in Section~\ref{sec:event_driven}: 
the detector's own confidence signal identifies precisely when 
the generative prior is needed, suppressing relighting in 
conditions where it would degrade performance.


\subsection{Dynamic: Cross-Controller Navigation}
\label{subsec:controller_comparison}

To validate that the generative sensor's semantic recovery directly drives safe, collision-avoidant maneuvers, we evaluate two underlying active controllers under Night ($\bullet$) conditions with continuous generative relighting. Table~\ref{tab:controller_comparison} presents the comparison between a baseline MPC (which strictly tracks static waypoints) and our MPC\_DLC (which leverages a dynamic lateral offset).

We discover that MPC\_DLC controller actively leverages dynamic reference tracking to reduce collisions rate from 28.8 to 22.0 per lap (${\sim}24\%$ reduction) compared to the baseline MPC. This establishes that the generative sensor's output can be seamlessly integrated into advanced predictive control schemes to execute evasive maneuvers, purposefully trading off an increase in average cross-track error (0.784~m vs.\ 0.539~m) for improved physical safety.

\begin{table}[ht]
\centering
\caption{Controller Comparison under Night ($\bullet$) conditions with Generative Relighting. Both controllers maintain 100\% lap completion by ingesting the generative perception manifold, but MPC\_DLC dynamically leverages tracking to actively reduce collisions by ${\sim}24\%$, naturally accepting an increase in cross-track error as it executes evasive maneuvers.}
\label{tab:controller_comparison}
\begin{threeparttable}
    \begin{tabular}{lcccc}
    \toprule
    \textbf{Controller} & \textbf{Collisions (\%)} & \textbf{CTE\tnote{1} (m)} & \textbf{Avg Latency} \\
    \midrule
    MPC (Relight)      & $28.8 \pm 3.7$ & $0.539 \pm 0.004$ & 62.0 ms \\
    MPC\_DLC (Relight) & $22.0 \pm 7.2$ & $0.784 \pm 0.153$ & 62.0 ms \\
    \bottomrule
    \end{tabular}
    
    \begin{tablenotes}
        \item[1] Cross-Track Error (CTE) represents the perpendicular distance from the center of the vehicle to the reference trajectory.
    \end{tablenotes}
\end{threeparttable}
\end{table}

\begin{figure}[t]
\centering
\setlength{\tabcolsep}{1pt}
\begin{tabular}{cccc}
 & \small\textbf{Day ($\circ$)} & \small\textbf{Sunset ($\oslash$)} & \small\textbf{Night ($\bullet$)} \\[1pt]

\rotatebox{90}{\hspace{3.0em}\scriptsize\textbf{Raw}} &
\includegraphics[width=0.32\columnwidth]{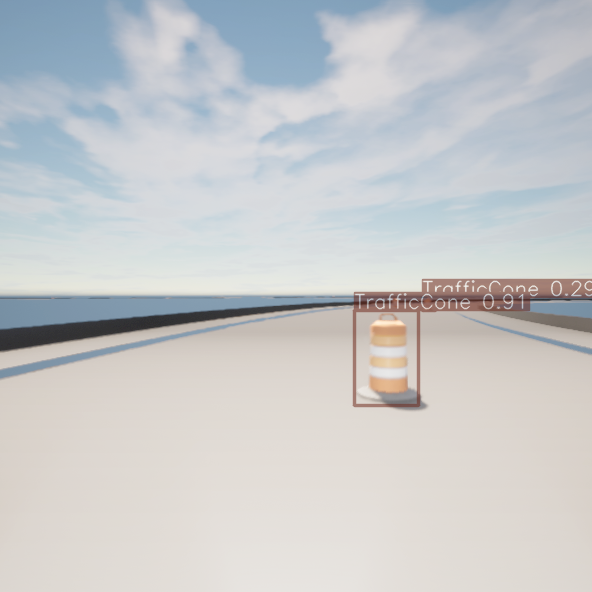} &
\includegraphics[width=0.32\columnwidth]{img/sunset_progress_020p_raw_frame_000450.png} &
\includegraphics[width=0.32\columnwidth]{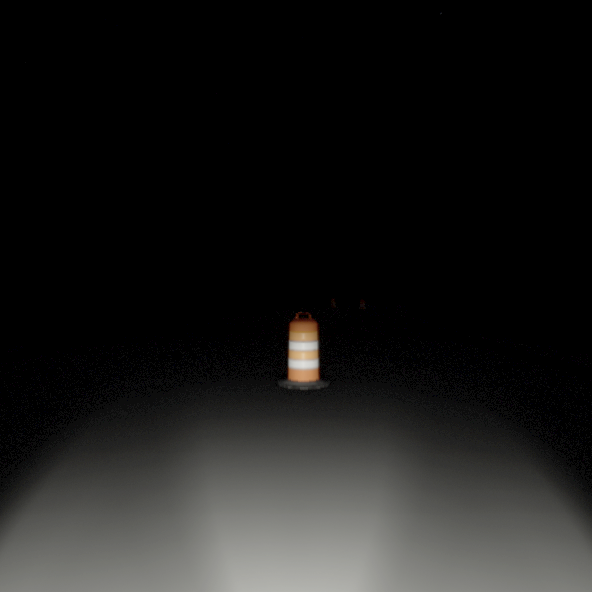} \\[1pt]

\rotatebox{90}{\hspace{3.0em}\scriptsize\textbf{Relit}} &
\includegraphics[width=0.32\columnwidth]{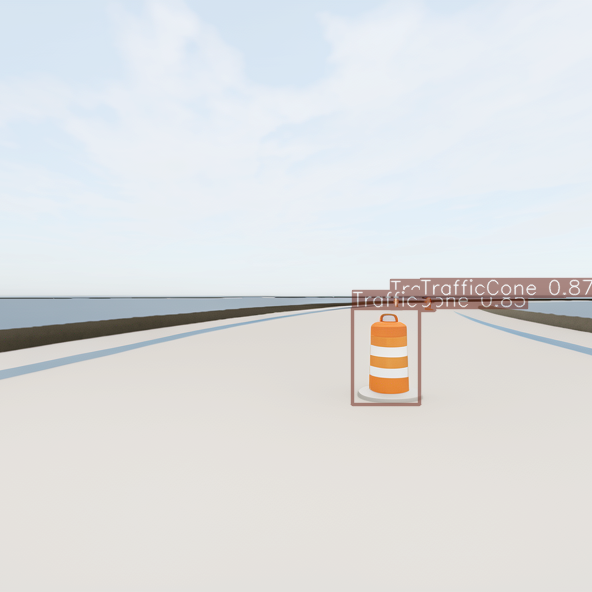} &
\includegraphics[width=0.32\columnwidth]{img/sunset_progress_020p_relit_frame_000450.png} &
\includegraphics[width=0.32\columnwidth]{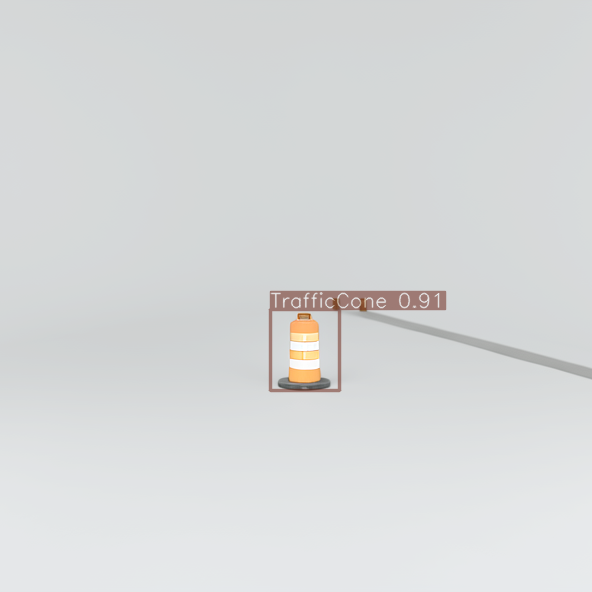} \\
\end{tabular}
\caption{Qualitative comparison of CARLA forward-camera views across lighting conditions, overlaid with YOLO bounding boxes. \textbf{Top:} Raw frames showing progressive perceptual degradation and subsequent detection failures in low light. \textbf{Bottom:} Generative outputs successfully restoring scene illumination and obstacle detection to near-daylight levels. While the generative process introduces structural hallucinations, critical obstacle visibility is maintained. The day condition (bottom-left) serves as an oracle reference.}
\label{fig:carla_qualitative}
\end{figure}

\subsection{Dynamic: Closed-Loop Navigation}
\label{subsec:safety_comparison}
To validate the necessity of the generative sensor for safe autonomous navigation, we evaluate the full system under strict, degraded lighting conditions. The MPC\_DLC controller is tasked with navigating the stochastically-obstructed Thunderhill track. Table~\ref{tab:safety_comparison} reports 5-lap collision statistics (mean $\pm$ std and collision rate) under both Sunset ($\oslash$) and Night ($\bullet$) degradation. 

Under both intermediate and extreme light degradation, the unenhanced baseline YOLO26 experiences severe performance collapse, resulting in massive collision rates ($32.75\%$ and $32.61\%$, respectively). Attempting to recover obstacle visibility through standard mathematical contrast stretching (CLAHE~\cite{pizer1990contrast}) proves completely insufficient, resulting in statistically equivalent or worse collision margins ($32.06\%$ and $34.39\%$ collision rates). Standard signal processing fails to restore the missing semantics.

In stark contrast, when the generative neural sensor is permitted to map the degraded physical environment back into the detector's optimal daylight distribution, navigation safety is robustly restored. At Sunset ($\oslash$), the generative sensor reduces the collision rate down to $21.44\%$, functionally mirroring the optimal Daylight Baseline ($22.47\%$). Under Night ($\bullet$) conditions, the generative system recovers the scene semantics to reduce the collision rate to $24.55\%$. This definitively proves that our generative prior reliably suppresses catastrophic vision failures and recovers near-daylight safety margins.

\begin{table}[ht]
\centering
\begin{tabularx}{\columnwidth}{X c c c}
\toprule
\textbf{Method} & \textbf{Collisions} & \textbf{Rate (\%)} & \textbf{Latency} \\
\midrule
Daytime (Oracle) & $20.4 \pm 2.4$ & $22.47 \pm 3.07$ & 1.7 ms \\
\midrule
Sunset (No Fix)  & $29.8 \pm 4.6$ & $32.75 \pm 4.45$ & 1.7 ms \\
Sunset + CLAHE   & $29.4 \pm 5.1$ & $32.06 \pm 3.71$ & 2.0 ms \\
Sunset + Gen. (Ours) & $\mathbf{19.4 \pm 2.9}$ & $\mathbf{21.44 \pm 3.54}$ & 62.0 ms \\
\midrule
Dark (No Fix)    & $29.6 \pm 3.6$ & $32.61 \pm 4.50$ & 1.7 ms \\
Dark + CLAHE     & $31.4 \pm 4.1$ & $34.39 \pm 3.64$ & 2.0 ms \\
Dark + Gen. (Ours) & $\mathbf{22.0 \pm 7.2}$ & $\mathbf{24.55 \pm 6.30}$ & 62.0 ms \\
\bottomrule
\end{tabularx}
\caption{Navigation Safety Under Low-Light Conditions. The generative neural sensor successfully suppresses collision events across varied lighting degradation severity.}
\vspace*{-1em}
\label{tab:safety_comparison}
\end{table}

\begin{table}[ht]
\centering
\begin{tabular}{l c c c c}
\toprule
\textbf{Mode} & \textbf{Collisions} & \textbf{Activated (\%)} & \textbf{Latency} & \textbf{GPU Util.} \\
\midrule
Continuous & $30.4 \pm 2.2$ & $100.0\%$ & 62.0 ms & $95.0\%$ \\
Event-Driven & $\textbf{23.4} \pm \textbf{2.1}$ & $\textbf{0.4\%}$ & \textbf{2.0 ms} & $\textbf{0.4\%}$ \\
\bottomrule
\end{tabular}
\caption{Event-Driven Triggering Evaluation at Sunset ($\oslash$). The hysteresis logic drastically reduces computational overhead---dropping GPU utilization from $95.0\%$ to $0.4\%$---while actively improving safety by limiting semantic hallucination strictly to frames experiencing genuine detection failure. ($\tau=0.3$) }
\label{tab:trigger_stats2}
\vspace*{-1em}
\end{table}

\subsection{Dynamic: Event-Driven Triggering Efficiency}
\label{subsec:event_driven_results}




\textcolor{black}{The continuous relighting evaluated in Section~\ref{subsec:safety_comparison} ran directly in the live CARLA control loop. To isolate the computational efficiency of the event-driven trigger without confounding the controller, we instead replayed our hysteresis logic ($\tau_{\text{low}}=0.3$, $\tau_{\text{high}}=0.5$) offline on synchronized confidence traces recorded from those closed-loop runs over 5 laps at Sunset ($\oslash$); full physical integration of the gating mechanism remains future work.}

\textcolor{black}{Table~\ref{tab:trigger_stats2} shows the resulting impact: since the baseline YOLO detector is not completely blind during transitional lighting, the trigger activated the generative sensor on only $0.4\%$ of frames, decoupling the neural overhead from the control loop and collapsing GPU utilization from $95.0\%$ to $0.4\%$ at a near-baseline $2.0$~ms average latency.}

\begin{figure}[ht]
\centering
\includegraphics[width=\linewidth]{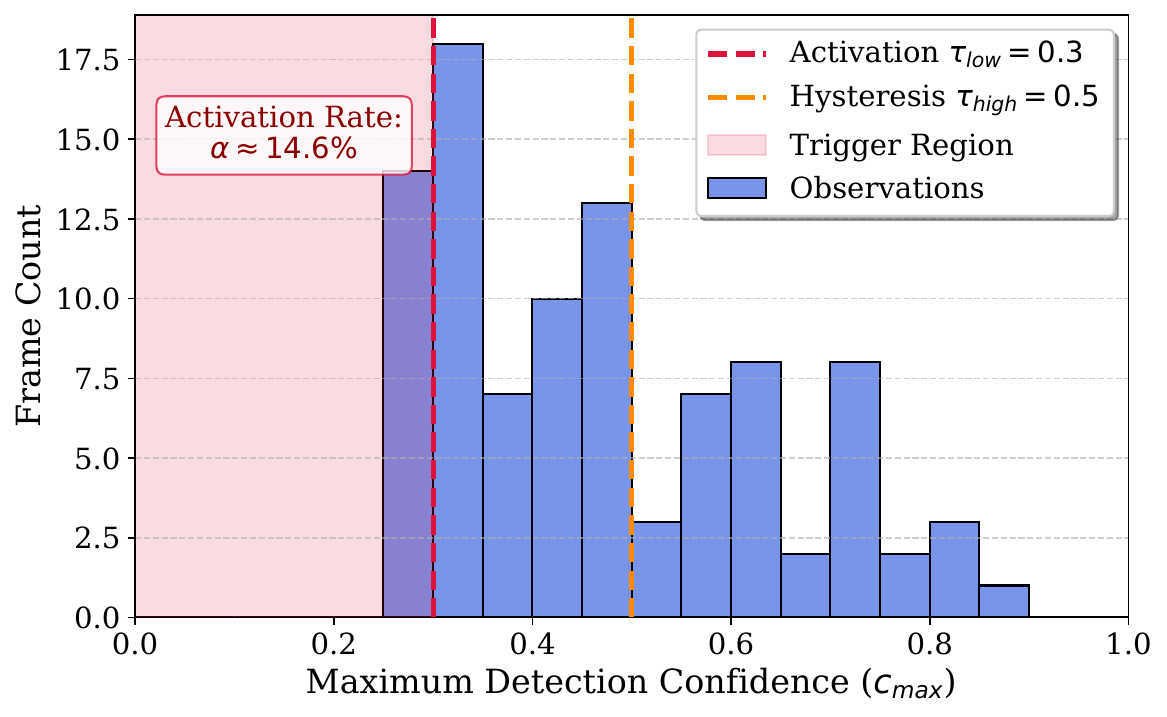}
\caption{Uncertainty-aware activation minimizes computational overhead. Distribution of maximum detection confidence $c_{\max}$ over a complete lap under \textbf{Night ($\bullet$, 22:00)} conditions. Vertical dashed lines mark $\tau_{\text{low}} = 0.3$ and $\tau_{\text{high}} = 0.5$. By treating generative enhancement as a sparse resource, the system selectively triggers the model, avoiding unnecessary latency and generative artifacts.}
\vspace*{-1em}
\label{fig:confidence_distribution}
\end{figure}

\subsection{Analysis of Generative Hallucination and Safety Risks}
\label{subsec:hallucination_analysis}
\textcolor{black}{Generative diffusion models prioritize visual plausibility over strict physical fidelity, risking semantic hallucination: our qualitative results occasionally show invented details, such as altered text fonts or unnatural textures (Fig.~\ref{fig:comparison_transpose}), rather than a faithful recovery of the true signal.}

\textcolor{black}{In safety-critical deployments, such hallucinations---shifting obstacle boundaries, removing faint cues, or inventing navigational features---pose real risks if ingested continuously by the control stack. Our Event-Driven Active Perception architecture mitigates this: gating the generative sensor behind the Perceptual Uncertainty Event threshold ($\tau_{\text{low}}$) defaults the system to the unenhanced camera stream whenever lighting is sufficient, confining hallucinations to cases where standard perception has already failed and keeping the diffusion prior a fail-safe rather than a continuous sensory replacement.}

\section{CONCLUSIONS}

We present the first system to deploy a large-scale generative
diffusion model as a real-time neural sensor within a live
robotic navigation stack, closing the deployment gap between
state-of-the-art image enhancement and real-time robotic
perception by combining a generative neural sensing module
with an uncertainty-aware activation mechanism in an
asynchronous architecture that treats generative enhancement
as a sparse computational resource.

Our evaluation yields three findings. First, on ExDark,
our relighting pipeline achieves 87.1\% mean AP---a 10.5
point improvement over MambaSFLNet
and the highest per-class AP across all 12 categories---
showing that a generative semantic prior outperforms
purpose-built discriminative enhancement on real-world
low-light detection. Second, on vKITTI2, our method raises
fog detection F1 from 0.681 to 0.855 (+0.174) while
MambaSFLNet degrades below the unenhanced baseline
(F1: 0.579), confirming that PSNR-optimized enhancement
suppresses the low-contrast edges that detectors rely on
under genuine semantic occlusion. Third, under conditions
where the baseline detector is already robust (Rain, Overcast,
Sunset), no enhancement method provides meaningful
benefit---motivating the confidence-triggered event-driven
design as the correct deployment strategy.
The asynchronous architecture achieves 16\,FPS at
62\,ms per-image latency, non-blocking to the 20\,Hz
control loop, with 100\% lap completion in CARLA.

Our evaluation is limited to single-class obstacles in CARLA, with the Delight LoRA
trained on a synthetic distribution. The event-driven trigger
was validated via offline replay rather than full closed-loop
integration, and the high GPU requirement currently limits
edge deployment.

Future work includes closing this loop with a direct
navigation-impact measurement, extending to multi-class
obstacles and long-duration real-world sequences, and
exploring model distillation to tighten real-time latency.
The triggering paradigm extends naturally to other
high-latency perception tasks, where uncertainty-aware
resource allocation can balance quality and computation cost.

\noindent

\section*{ACKNOWLEDGMENT}
This project is supported in part by Dr. Barry Mersky and Capital One E-Nnovate Endowed Professorships and UMD-ARL Cooperate Agreement.










\balance

\bibliographystyle{IEEEtran}
\bibliography{references} 


\end{document}